# Approche Hybride pour la translitération de l'arabizi algérien : une étude préliminaire

Imane Guellil [1, 2] Faical Azouaou[1] Fodil Benali [1] ala-eddine Hachani[1] Houda Saadane[3]
(1) Ecole nationale Supérieure d'Informatique, BP 68M, 16309, Oued-Smar, Alger,
(2) Ecole Supérieure des Sciences Appliquées, Alger, Algérie, http://www.essa-alger.dz
(3) GEOLSemantics, 12 Avenue Raspail, 94250 Gentilly, France,
i_guellil@esi.dz, f_azouaou@esi.dz, df_benali@esi.dz, da_hachani@esi.dz,
houda.saadane@geolsemantics.com

## RESUME

Dans cet article, nous présentons une approche hybride pour la translitération de l'arabizi algérien. Nous avons élaboré un ensemble de règles permettant le passage de l'arabizi vers l'arabe. Á partir de ces règles nous générons un ensemble de candidats pour la translitération de chaque mot en arabizi vers l'arabe, et un parmi ces candidats sera ensuite identifié et extrait comme le meilleur candidat. Cette approche a été expérimentée en utilisant trois corpus de tests. Les résultats obtenus montrent une amélioration du score de précision qui était pour le meilleur des cas de l'ordre de 75,11%. Ces résultats ont aussi permis de vérifier que notre approche est très compétitive par rapport aux travaux traitant de la translitération de l'arabizi en général.

## ABSTRACT

**A hybrid approach for the transliteration of Algerian Arabizi: A primary study**
In this paper, we present a hybrid approach for the transliteration of the Algerian Arabizi. We define a set of rules for the passage from Arabizi to Arabic. Through these rules, we generate a set of candidates for the transliteration of each Arabizi word into arabic. Then, we extract the best candidate. This approach was evaluated by using three test corpora, and the obtained results show an improvement of the precision score which is equal to 75.11% for the best result. These results allow us to verify that our approach is very competitive comparing to others works that treat Arabizi transliteration in general.

MOTS-CLES : arabizi, Dialecte algérien, arabizi algérien, Translitération.

# 1 Introduction

*L'arabizi est une orthographe spontanée utilisée pour s'exprimer en dialecte arabe combinant lettres latines, chiffres et autres symboles tels que les signes de ponctuations (Al-Badrashiny et al., 2014; Guellil, AZOUAOU, 2017)* . L'arabizi est généralement utilisé par les locuteurs arabes pour les échanges dans les réseaux sociaux, les chats voir les SMS. Néanmoins, comme la plupart des outils de traitement des dialectes arabes, et du dialecte algérien (DALG) en particulier, sont dédiés aux messages écrits en lettres arabes, par conséquent l'arabizi ne peut donc être traité comme tel. La plupart des recherches s'oriente vers la transformation de l'arabizi vers l'arabe. Cette transformation

est nommée *la translitération*. La translitération est un processus de passage d'un texte écrit en un script ou alphabet donné vers un autre (Guellil et al., 2017a; Guellil et al., 2017b; Josan, Lehal, 2010; Kaur, Singh, 2014). La translitération de l'arabizi vers l'arabe fait cependant face à un ensemble de problématiques :

**1) Traitement des voyelles :** les voyelles (a, i, o, u, e, y) peuvent être remplacées par les différentes lettres arabes (ة,و,ي,ا,أ) ou encore par aucune lettre. Cela dépend de leurs emplacements dans le mot.

**2) L'ambiguïté entre plusieurs lettres :** une lettre arabizi peut correspondre à plusieurs lettres arabes. Par exemple, la lettre 't' peut correspondre aux deux lettres arabes 'ت' t et 'ط' T.

**3) L'ambiguïté reliée au contexte :** dans certains cas, plusieurs translitérations peuvent correspondre au même mot. Par exemple le mot 'matar' pourrait être translitéré en 'مطر' *maTar*[1] 'la pluie' ou encore 'مطار' *maTaAr* 'aéroport' (Al-Badrashiny et al., 2014).

**4) Ambiguïté reliée au code switching[2] :** certains mots d'autre langues tels que le français ou l'anglais peuvent être pris pour des mots en arabizi. Par exemple le mot 'men' en anglais.

Afin d'adresser ces problématiques, nous présentons dans cet article une approche hybride pour la translitération de l'arabizi algérien. Dans la suite du présent article, nous présentons dans la section 2 une synthèse de l'état de l'art liée à la problématique de la translittération, puis nous mettons en avant dans la section 3 l'originalité de notre approche par rapport aux travaux étudiés. Ensuite nous décrivons dans la section 4 notre approche proposée. Quant à la section 5, elle sera consacrée à la présentation des expérimentations menées ainsi que les résultats obtenus. Enfin, dans la section 6 nous concluons cet article avec une présentation de nos travaux futurs.

## 2 État de l'art

Le problème de la translittération a suscité l'intérêt des spécialistes dans plusieurs langues. Cet intérêt récent est lié au développement croissant de l'utilisation d'Internet (Saâdane, Semmar, 2012) Trois approches sont couramment évoquées dans la littérature pour réaliser la translittération: 1) Á base de règles. 2) Statistiques et 3) Hybrides combinant les deux précédentes ( Guellil et al., 2017; Kaur, Singh, 2014). Habash, et al.,(2007), ont proposé un ensemble de règles permettant le passage de l'arabizi vers l'arabe. Ils ont signalé un nombre d'exceptions et de défis reliés aux traitements des voyelles. Rosca&Breuel (2016) ont abordé l'approche statistique où ils ont présenté un modèle basé sur les réseaux de neurones pour effectuer la translitération entre plusieurs paires de langues dont l'arabe et l'anglais. L'approche hybride est utilisée dans les travaux de Al-Badrashiny et al., (2014); Darwish (2013); Guellil et al., (2017a); Guellil et al., (2017b);Mayet al., (2014); Saâdane et al., (2017); Saâdane et al., (2013); van der Wees et al., (2016). Tous ces travaux suivent la même idée générale, à savoir générer un ensemble de possibilités de translittération, appelés candidats, pour ensuite déterminer le meilleur candidat à l'aide d'un modèle de translitération ou autre. Pour ce faire, Darwish (2013) construit manuellement un ensemble contenant 3452 mots arabizi (extrait de Twitter) translitéré vers l'arabe. Une partie de ce corpus arabizi-arabe a été utilisée dans le travail de (Al-Badrashiny et al., 2014). Les auteurs de ce travail font cependant appel à un automate d'état fini pour le passage de l'arabizi vers l'arabe. Dans les travaux de(May et al., 2014; van der Wees et al., 2016), les auteurs analysent également les résultats de la traduction après la phase de translitération. Enfin, les travaux de (Guellil et al., 2017; Saâdane et al., 2017; Saâdaneet al., 2013)

---

[1]Translittération arabe présentée dans schemeHabash-Soudi-Buckwalter (HSB) (Habash et al., 2007)

[2] Code switching: Présence de plusieurs langues ou dialectes au sein du même message

constituent les uniques références que nous avons recensées sur l'arabizi algérien. Guellil et al., (2017) ont développé un algorithme basé sur les règles pour la translitération en s'appuyant sur un corpus arabe qu'ils ont translitéré en arabizi. Ces auteurs ajoutent ensuite la notion de traduction automatique de l'arabizi après la translitération (Guellil et al., 2017). Saâdane et al., (2017); Saâdane et al., (2013) ont translitéré le texte arabizi en arabe en utilisant un automate d'état fini. Les résultats générés sont normalisés en suivant la convention de transcription nommée *CODA* "Conventional Orthography for Dialectal Arabic (Saâdane, Habash, 2015). Ensuite les nouveaux résultats sont filtrés en utilisant un analyseur morphologique de l'arabe. Nous signalons cependant que ces travaux se focalisent plus sur l'identification de l'origine dialectale.

## 3  Positionnement de notre approche par rapport aux travaux étudiés

Nous présentons dans cet article une approche de translitération de l'Arabizi Algérien vers l'arabe. Pour définir cette approche, nous nous sommes appuyés sur plusieurs travaux comme suit :

1) Nous avons utilisé une table de passage de l'arabizi vers l'arabe similaire à celle présentée dans les travaux par van der Wees et al., (2016), qui est extraite de Wikipédia[3]. Cependant, nous avons défini en plus des règles de passage ainsi que nos propres règles dédiées au traitement de l'Arabizi Algérien, notamment les différentes ambiguïtés de translitération. Pour ce faire, nous nous sommes inspirés des travaux de (Guellil et al., 2017) décrivant les caractéristiques de l'Arabizi Algérien.

2) Les travaux de Al-Badrashiny et al., (2014); Darwish (2013); Guellilet al., (2017a); Guellil, et al., (2017b); May et al., (2014); van der Wees et al.,(2016) génèrent un ensemble de candidats pour la translitération d'un mot arabizi en arabe. Par exemple, les candidats générés des travaux de ces auteurs pour le mot '3afsa' (une astuce) sont : عافسة '*çaAfsaħ* , عافسا '*çaAfsaA*, عافزة '*çaAfzaħ*, عافزا'*çaAfzaA*,, etc. Néanmoins nous ne trouvons aucun candidat sous la forme de عفسة 'qui est la translitération correcte de ce mot. La valeur ajoutée de notre travail est que notre algorithme est capable de générer de tels candidats en remplaçant les voyelles par le caractère vide (NULL).

3) Les travaux de Darwish ( 2013); Guellil et al., (2017a); Guellilet al., (2017b) qui font appel à un modèle de translitération pour déterminer le meilleur candidat. Cependant, ces approches dépendent d'un corpus parallèle, correspondant à la translitération d'un ensemble de messages de l'arabizi vers l'arabe, en plus du coût élevé de la réalisation de ces même corpus. Pour notre part, au lieu de construire des corpus parallèles arabizi-arabe., nous appliquons un modèle de translitération sur un corpus Arabe assez volumineux extrait des médias sociaux (par nos soin) et comparant ainsi les résultats obtenus.

## 4  Approche de translitération de l'arabizi algérien vers l'arabe

Notre approche se compose de quatre phases importantes : 1) Extraction et prétraitement du corpus Arabe et des messages. 2) Proposition et application des règles pour l'arabizi algérien. 3) Génération des différents candidats et 4) Extraction du meilleur candidat. Nous présentons dans la Figure 1 l'architecture générale de notre approche.

---

[3]https://en.wikipedia.org/wiki/Arabic_chat_alphabet

## 4.1 Extraction et prétraitement du corpus Arabe et du message arabizi

Nous commençons tout d'abord par l'extraction d'un corpus Arabe issu des réseaux sociaux et contenant un ensemble de commentaires de locuteurs Algériens. Pour ce faire, nous avons ciblé un ensemble de pages très populaires en Algérie comme la page **Ooredoo**[4] (opérateur téléphonique). Après l'extraction de ce corpus, nous nous sommes focalisés sur les messages écrits uniquement en caractères arabes. Nous supprimons ensuite l'exagération (par exemple, le mot hiaaaaati–*ma vie*-est transformé en hiati) de ce corpus et nous remplaçons les différents caractères arabes par leurs Unicodes. Nous procédons ensuite aux mêmes prétraitements sur le message arabizi

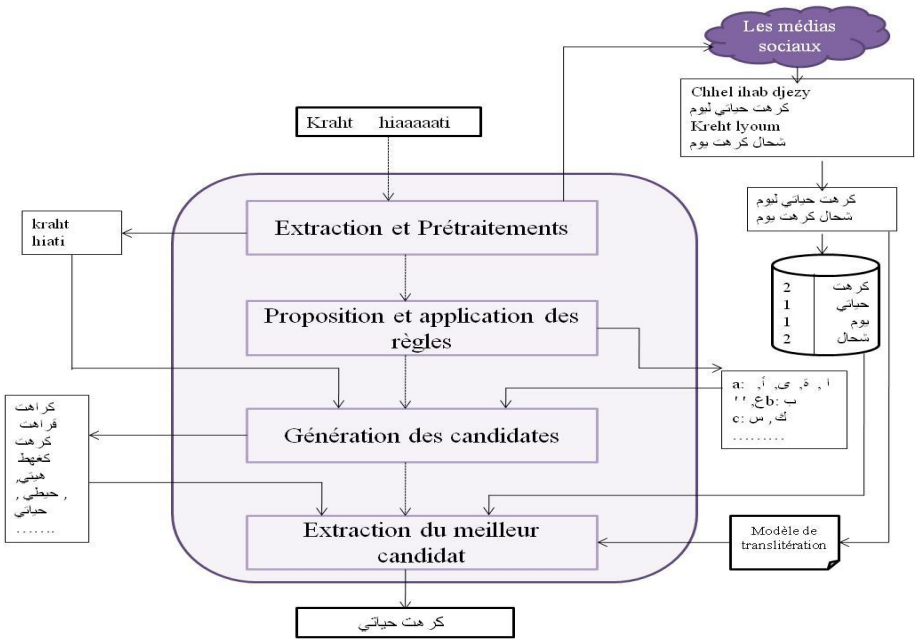

Figure *1*: Architecture générale de notre approche

## 4.2 Proposition et application des règles pour l'arabizi algérien

Nous présentons dans le tableau 1, les différentes possibilités de remplacement de chaque lettre en arabizi algérien. En plus de ce tableau, nous définissons un ensemble de règles de passage de l'arabizi vers l'arabe. Par exemple : la voyelle 'a' est remplacée par la lettre 'أ' au dédut. Au milieu du mot, cette lettre peut être remplacée soit par le caractère Arabe 'ا' ou encore par le caractère vide.

## 4.3 Génération des différents candidats

En appliquant les différents remplacements du tableau 1 ainsi que des différentes règles élaborées, chaque mot arabizi donne naissance à plusieurs mots en arabe. Pour illustrer cela, reprenons l'exemple des deux mots « kraht » et « hiati ». Le mot « kraht » donne lieu à 32 candidats. Parmi ces derniers, nous citons: كراهت *krAht*, قراهت *qrAht*, كرهت *krht*, كغهط *kyhT*, قرحت *qrHt*, etc. Le mot correctement translitéré étant «كرهت» *krht*. Quant au mot « hiati », nous avons 16 candidats parmi lesquels nous citons: هياتي *hyAty*, هيتي *hyty*, حيطي *HyTy* , حياتي *HyAty*, حيتي *Hyty*, etc. Le mot correctement translitéré étant «حياتي» *HyAty*».

---

[4]https://fr-fr.facebook.com/OoredooDZ/

| Lettre en Arabizi | Lettre en arabe | Lettre en Arabizi | Lettre en Arabe | Lettre en Arabizi | Lettre en Arabe |
|---|---|---|---|---|---|
| A | ", ا , ق, ى, أ, ع | k | ك, ق | U | ", و , أ |
| B | ب | l | ل | V | ف |
| C | س , ك | m | م | W | و |
| D | د, ض, ظ | n | ن | X | كس |
| E | ",ا | o | و ,",أ | Y | ", إ, ي |
| F | ف | p | ب | Z | ز |
| G | ق | q | ك | 7 | ح |
| H | ه , ح | r | ر,غ | 5 | خ |
| I | ",ي | s | س , ص | 3 | ع |
| J | ج | t | ت ط | 9 | ق |

Table 1 : Lettres de passage de l'arabizi vers l'Arabe

## 4.4 Extraction du meilleur candidat à la translitération

Pour extraire le meilleur candidat de translitération d'un mot arabizi, nous réalisons les deux étapes suivantes : 1) réalisé une recherche simple de chaque candidat au sein de notre corpus arabe, et 2) effectuer une recherche basée sur un modèle de langue appliqué sur notre corpus arabe. Au cours de la recherche simple, nous recherchons chaque candidat au sein de notre ensemble de mots afin de récupérer le nombre d'occurrence de chaque candidat. Pour la recherche basée sur un modèle de translitération, nous recherchons chaque candidat au sein de notre modèle en extrayant la probabilité de chacune. Nous retournons ensuite le candidat ayant la probabilité la plus élevée.

# 5 Expérimentations et résultats

## 5.1 Le corpus arabe utilisé

Pour la création de nos ressources linguistiques (corpus), nous avons tout d'abord ciblé 133 pages Facebook Algérienne dont : *Ooreedo, Djezzy, HamoudBoualem,* etc. Ces pages concernent les opérateurs téléphoniques, des producteurs de limonade ou de jus de fruit, la presse, etc.

Nous avons également découpé le corpus PADIC (Meftouh, Harrat, Jamoussi, Abbas, & Smaili, 2015) en un ensemble de termes et construit un dictionnaire du dialecte algérien, que nous avons pu récolter grâce à la traduction d'un dictionnaire anglais en faisant appel à l'API glosbe[5]. A l'aide de l'API de Facebook (RestFB[6]), et grâce à la fonction « *search*» de RestFB, nous extrayons l'ensemble des posts et commentaires relatifs aux pages et différents dictionnaires. Par ailleurs, nous signalons que les sources censées alimenter notre corpus concernent divers sujets : politique, sport, économie, etc. Notre corpus contient **3 668 575** messages et ce après prétraitement et extraction des messages écrits en caractère arabe.

---

[5] https://glosbe.com/
[6] http://restfb.com/

## 5.2 Les corpus de test utilisés

Nous avons fait appel à trois corpus de tests :
1) Test_300: Contenant 300 messages (2005 mots) extrait de Facebook. Ces messages font partie du corpus contenant **3 668 575** que nous avons extrait précédemment.
2) Test_50 : Contenant 50 messages (527 mots) messages extrait du corpus de Cotterell, et al.,(2014). Ce corpus étant le seul corpus de l'arabizi algérien en libre accès au sein de la communauté de recherche. Ce corpus a déjà été utilisé comme corpus de test au sein des travaux dans (Imane Guellil, Azouaou, Abbas, et al., 2017).
3) Test_200 : Contenant 200 messages (933 mots) extraient du corpus Parallèle PADIC (Meftouh et al., 2015). Néanmoins ce corpus est en caractère arabe mais il a été translitéré vers l'arabizi dans les travaux de Imane Guellil et al., (2017).

## 5.3 Les résultats expérimentaux

Pour nos expérimentations, nous découpons notre corpus arabe en plusieurs parties. Chaque découpage a été utilisé pour entraîner un modèle de translitération et pour faire la recherche simple basée sur notre algorithme. Nous menons ainsi plusieurs expérimentations où nous utilisons respectivement : 1% (36 682 messages/ 63 269 mots), 5%(183 413 messages/ 177 722 mots), 10% (366 827 messages/ 268 751 mots), 25% (917 068 messages/ 454 817 mots), 50% (1 834 137 messages/ 658 738 mots), 75% (2 751 205 messages/ 810 611 mots) et finalement 100% (3 668 275 messages/ 930 462 mots) de notre corpus arabe. Pour le modèle de translitération nous faisons appel à l'implémentation JAVA du modèle KenLM[7]. Nous testons notre algorithme pour différents n-gramme (avec n allant de 2 à 5). Nous nous sommes rendus compte que nous obtenions pratiquement les mêmes résultats à chaque fois, car la recherche du meilleur candidat se fait en 1-gramme uniquement. De ce fait, nous avons décidé d'utiliser un modèle de translitération avec 2-gramme. Nous présentons dans le tableau 2, l'ensemble des résultats de translitération de l'arabizi algérien en se basant sur les deux approches décrites précédemment (recherche simple et modèle de translitération avec kenLM). Nous calculons pour chaque corpus de test l'accuracy qui est défini comme suit :
**l'Acurracy= nombre de mot correctement translitéré/ nombre total de mot dans le corpus**

| Approche | Corpus | 1% | 5% | 10% | 25% | 50% | 75% | 100% |
|---|---|---|---|---|---|---|---|---|
| Recherche simple | Test_50 | 67.36 | 70.02 | 72.49 | 73.24 | 74.19 | 74.57 | **74.76** |
| | Test_200 | 67.31 | 69.74 | 69.67 | 70.95 | 71.06 | 71.28 | 72.03 |
| | Test_300 | 68.79 | 72.02 | 72.97 | 73.72 | 74.16 | 74.71 | **75.11** |
| KenLM | Test_50 | 66.79 | 70.21 | 71.54 | 72.87 | 72.49 | 72.30 | 72.67 |
| | Test_200 | 64.84 | 66.88 | 67.85 | 68.38 | 69.45 | 69.24 | 69.34 |
| | Test_300 | 68.33 | 70.57 | 71.73 | 72.37 | 72.76 | 73.07 | 73.27 |

Table 2 : Résultats de translitération de l'Arabizi algérien

---

[7] https://github.com/jbaiter/kenlm-java

## 5.4 Analyse des résultats et des cas d'erreurs

D'après le tableau 2, nous constatons que la taille du corpus arabe influe sur les résultats obtenus. Plus ce corpus est volumineux, meilleurs sont les résultats. Concernant les corpus de test, nous avons utilisé trois corpus contenant respectivement (50, 200 et 300) messages. Nous constatons une amélioration remarquable au sein du corpus Test_50 où la précision dans Imane Guellil et al., (2017) est à 45.35% et dans notre cas elle atteint 74,76% dans le cas de la recherche simple et où le corpus arabe utilisé est complet (c'est-à-dire 100%). Nous obtenons une précision égale à 75,11% dans le cas de notre corpus Test_300, qui représente le meilleur résultat obtenu, ce qui est compréhensible vu que notre approche est basée sur la translitération des messages extraient des médias sociaux et que la translitération est faite de l'arabizi vers l'arabe et non pas l'inverse comme cela est fait dans dans Imane Guellil et al., (2017).

Néanmoins en analysant le corpus translitéré, nous avons identifié les erreurs suivantes : 1) Omission de certaines voyelles où elles devraient apparaître. Par exemple : le mot 'bik' est translitéré en 'بك' au lieu de 'بيك'. 2) Présence de certaines voyelles alors qu'elles ne devraient pas apparaître. Par exemple le mot 'bark' est translitéré en 'بارك' au lieu de 'برك', le mot 'lawel' est translitéré en 'لاول' au lieu de 'لول'. 3) Dans certains cas deux translitérations sont correctes, tout dépend du contexte et du sens de la phrase. Par exemple, le mot 'raht' pourrait être translitéré en 'رحت' ou en 'راحت', ou encore le mot 'djabat' qui pourrait être translitéré en 'جبت' ou en 'جابت' et ce tout dépend du sens de la phrase. 4) Des erreurs reliées aux mots puisant leurs signification du français et donc non reconnu par notre corpus dans la plupart des cas. Par exemple, le mot 'lafichage' est translitéré en 'لافيشاج' au lieu de 'لفيشاج' et le mot 'elsemastar' est translitéré en 'السوماستر' au lieu de 'السمستر'.Toutes ces erreurs sont causé par deux principales raisons : 1) La non prise en considération du contexte du mot dans la phrase. 2) Au non traitement des mots ayant comme signification une langue étrangère (principalement le français).

# 6   Conclusion et perspectives

Dans cet article, nous avons présenté une approche hybride de translitération de l'arabizi algérien vers l'arabe. Cette approche est basée sur la combinaison entre règles et modèles statistiques pour déterminer le meilleur candidat répondant à la translitération d'un mot en arabizi. Notre approche pourrait cependant être améliorée en y intégrant les points suivants :

- L'utilisation d'un corpus plus volumineux pour améliorer les résultats obtenus car nous avons constaté que la taille du corpus influe sur les résultats renvoyés. Il serait également intéressant de se pencher sur le niveau caractère.
- Mis en place d'une approche qui formerait des candidats contenant des n-gramme et non seulement des 1-gramme. Ceci nous aidera à situer le mot au sein de la phrase et non pas le traiter comme entité seule.
- Traiter le cas des mots étranger, par exemple, les mots français.
- Utiliser cette approche pour générer un corpus parallèle arabizi-arabe de manière semi-supervisée. Ce corpus pourrait être utilisé pour générer un modèle statistique.

Enfin, nous signalons que les corpus développés dans le cadre de cette étude seront bientôt mis à la disposition de la communauté scientifique.